\documentclass{article}

\usepackage{arxiv}

\usepackage[utf8]{inputenc} 
\usepackage[T1]{fontenc}    
\usepackage{hyperref}       
\usepackage{url}            
\usepackage{booktabs}       
\usepackage{amsfonts}       
\usepackage{nicefrac}       
\usepackage{microtype}      
\usepackage{lipsum}
\usepackage{graphicx}
\usepackage{amsmath}
\usepackage{amssymb}
\usepackage{xcolor}

\usepackage[ruled,vlined]{algorithm2e}
\usepackage{multirow}

\usepackage{xcolor}  
\usepackage{wrapfig}
\usepackage{duckuments}
\usepackage{CJKutf8}
\usepackage{overpic}
\usepackage{pifont}
\usepackage{bbding}
\usepackage{comment}
\usepackage{float}

\definecolor{MyDarkBlue}{rgb}{0,0.5,1}
\definecolor{MyDarkGreen}{rgb}{0.02,0.6,0.02}
\definecolor{MyDarkRed}{rgb}{0.8,0.02,0.02}
\definecolor{MyDarkOrange}{rgb}{0.40,0.2,0.02}
\definecolor{MyYellow}{rgb}{1,0.55,0}
\definecolor{MyPurple}{RGB}{111,0,255}
\definecolor{MyRed}{rgb}{1.0,0.0,0.0}
\definecolor{MyGold}{rgb}{0.75,0.6,0.12}
\definecolor{MyDarkgray}{rgb}{0.66, 0.66, 0.66}
\definecolor{default}{RGB}{0,0,0}

\newcommand\eg{\textit{e.g., }}
\newcommand\ie{\textit{i.e., }}

\newcommand{\model}{BendTwin} 
\renewcommand{\eqref}[1]{Eq.~(\ref{#1})} %

\graphicspath{ {./images/} }

\title{BendTwin: Robust Dense-to-Sparse Physical Reconstruction with Bending-Aware Differentiable Spring–Mass Models}

\author{
  Yixiong Jing\thanks{Equal contribution.} \\
  University of Cambridge\\
  \texttt{yj401@cam.ac.uk}
  \And
  Qi Wang\footnotemark[1] \\
 Institute of Automation, Chinese Academy of Sciences\\
  \texttt{qi.wang@ia.ac.cn}
  \AND
  Lin Chen \\
  Northwestern Polytechnical University\\
  \texttt{npuchenlin@foxmail.com}  
  \And
  Junwei Jiang \\
  The Hong Kong Polytechnic University\\
  \texttt{junwei.jiang97@outlook.com}  
  \And
  Guangming Wang\thanks{Corresponding author.} \\
  University of Cambridge\\
  \texttt{gw462@cam.ac.uk}
  \AND
  Haibing Wu \\
  University of Cambridge\\
  \texttt{hw657@cam.ac.uk}
  \And
  Olaf Wysocki \\
  University of Cambridge\\
  \texttt{okw24@cam.ac.uk}
  \And
  Wanli Ma \\
  University of Cambridge\\
  \texttt{wm369@cam.ac.uk}
  \And
  Brian Sheil \\
  University of Cambridge\\
  \texttt{bbs24@cam.ac.uk}
}

\begin{document}
\maketitle

\begin{abstract}
Reconstructing objects with mechanical properties from video observations enables physically consistent dynamic prediction, benefiting robotics planning and interaction. Existing spring--mass based physical driven reconstruction approaches offer efficient and differentiable physical reconstruction, but they typically rely on axial springs alone. Such formulations oversimplify the underlying structural mechanics and can become mechanically under-constrained when the physical graph is coarsened, limiting their ability to preserve stable local deformation. We present \model{}, a bending-aware differentiable spring--mass framework for video-based reconstruction and future prediction of deformable objects. \model{} introduces bending stiffness and damping over local surface triplets, penalizing deviations from rest angles and regularizing higher-order deformation. These bending constraints improve mechanical stability while preserving the simplicity of spring--mass system. Experiments show that \model{} consistently outperforms the axial-only PhysTwin baseline. Ablation studies further demonstrate that the bending constraints maintain system stability across different downsampling ratios and consistently improve upon the original PhysTwin formulation. Overall, \model{} provides an effective approach for constructing mechanically faithful digital twins from sparse-view RGB-D videos.
\end{abstract}


\section{Introduction}
\label{sec:introduction}
Creating digital twins of real-world objects is essential for predicting how objects move and deform under interaction. Accurate simulation of future dynamic states from a physically consistent digital twin enables a wide range of downstream applications, including robot manipulation planning~\cite{pmlr-v270-li25c, jangir2025robotarena, 2025arXiv251104665Z}, data generation for policy learning~\cite{chebotar2019closing, muratore2022neural, pmlr-v139-mora21a}, and physically grounded rollouts within vision-based world models~\cite{guo2025ctrl}. To obtain such digital twins, it is critical to infer physically plausible models directly from observations such as RGB-D videos, reducing manual modeling effort and helping bridge the gap between simulated dynamics and real-world behavior.

However, precise reconstruction of the physical model of a real object remains non-trivial. Previous graphics and engineering pipelines rely on carefully designed constitutive models and discretizations~\cite{sifakis2012fem, muller2007position, stomakhin2013material} but typically assume that the geometry and mechanical properties of objects are either known a priori or identified through controlled experiments. While these approaches can be accurate, they are often expensive and difficult to generalize.

\begin{figure}[tb]
  \centering
  \includegraphics[width=1.0\linewidth]{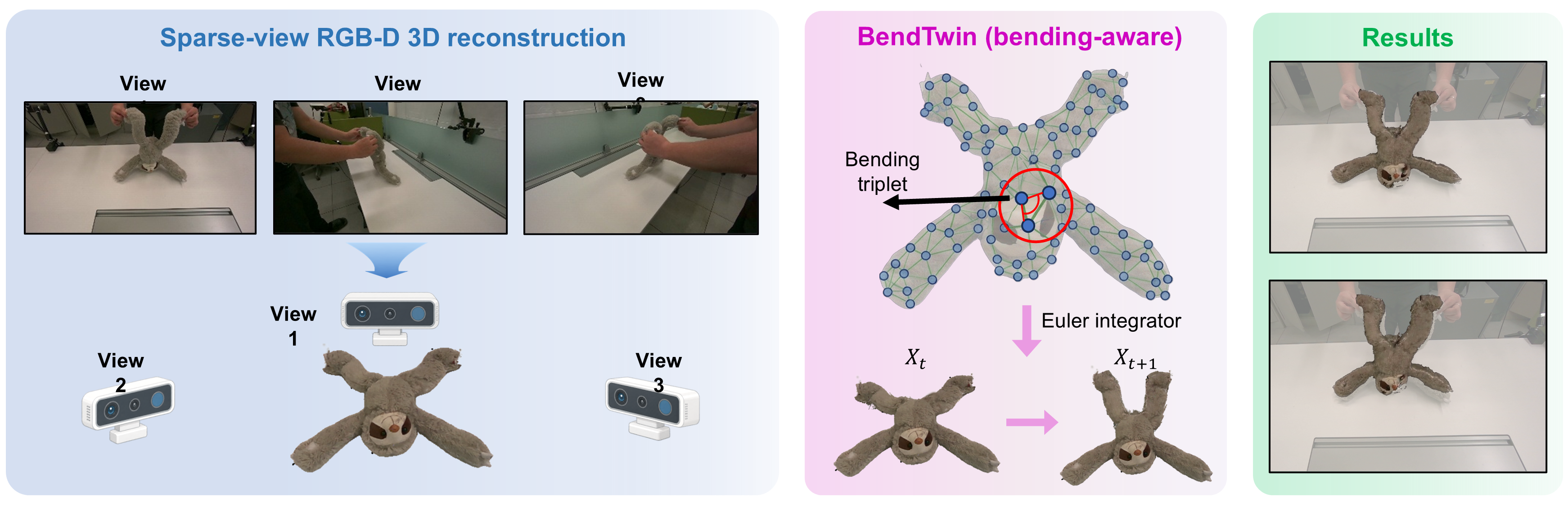}
  \caption{Overview of \model{}, which reconstructs deformable objects from sparse-view RGB-D videos and models them using a bending-aware surface spring--mass system for differentiable reconstruction \& simulation, and prediction.}
  \label{fig:framework}
\end{figure}

Recent advances in 3D reconstruction~\cite{kerbl20233d, mildenhall2021nerf} and generative models~\cite{11094803, xiang2025trellis2} enable detailed geometric modeling of real-world objects, and when combined with classical continuum mechanics simulators, can produce visually convincing deformations~\cite{zhang2024physdreamer, xie2024physgaussian}. However, material properties in prior work are typically manually assigned or generatively produced, and therefore do not necessarily reflect the true dynamics of the corresponding objects, limiting physical fidelity. Differentiable physics~\cite{qiao2021efficient, qiao2020scalable, du2021diffpd} addresses this limitation by enabling direct optimization of unknown physical parameters in governing equations through matching simulated dynamics to observations. PhysTwin~\cite{jiang2025phystwin} leverages this paradigm by reconstructing object geometry from sparse-view RGB-D videos using 3D Gaussian Splatting~\cite{kerbl20233d} and TRELLIS~\cite{11094803}, and optimizing a differentiable spring--mass system with learnable mechanical and contact parameters while enforcing visual consistency via Gaussian rendering.

However, PhysTwin~\cite{jiang2025phystwin} primarily relies on axial springs and damping. While effective for capturing stretching-dominated motion, axial-only formulations constrain only pairwise distances and do not explicitly regulate local angular deformation. As the physical graph (\eg the node counts for PhysTwin~\cite{jiang2025phystwin}) becomes sparse with fewer axial connections, the limited mechanical expressiveness of axial-only formulations can leave the system under-constrained. As a result, the model can struggle to preserve stable local deformation.

To address this limitation, we introduce \textbf{\model{}}, a bending-aware differentiable spring--mass framework illustrated in Fig.~\ref{fig:framework}. \model{} augments the conventional axial spring formulation with explicit bending stiffness and damping defined over local surface triplets. These bending constraints complement axial length constraints by penalizing deviations from rest angles, enabling the model to better preserve local shape and resist unstable deformation modes.
As a result, \model{} remains mechanically stable even as the spring--mass graph is coarsened to far fewer nodes and connections, while preserving the
simplicity of explicit spring--mass dynamics.

Our experiments demonstrate consistent improvements over the axial-only PhysTwin baseline in reconstruction \& re-simulation and future prediction. Ablation studies further show that explicit bending improves mechanical fidelity and maintains stable behavior as the physical graph becomes increasingly sparse under different downsampling ratios. In summary, our contributions are:

\begin{itemize}
    \item We propose \model{}, a bending-aware differentiable spring--mass framework that augments axial interactions with explicit bending stiffness and damping over local surface triplets, improving mechanical expressiveness for video-based physical reconstruction \& simulation, and prediction.

    \item We show that explicit bending constraints stabilize the spring--mass system under
    different downsampling ratios, allowing \model{} to retain accuracy with far fewer nodes where axial-only formulations become under-constrained.
\end{itemize}
\section{\model{}}
\label{sec:method}

\begin{figure}[tb]
  \centering
  \includegraphics[width=1.0\linewidth]{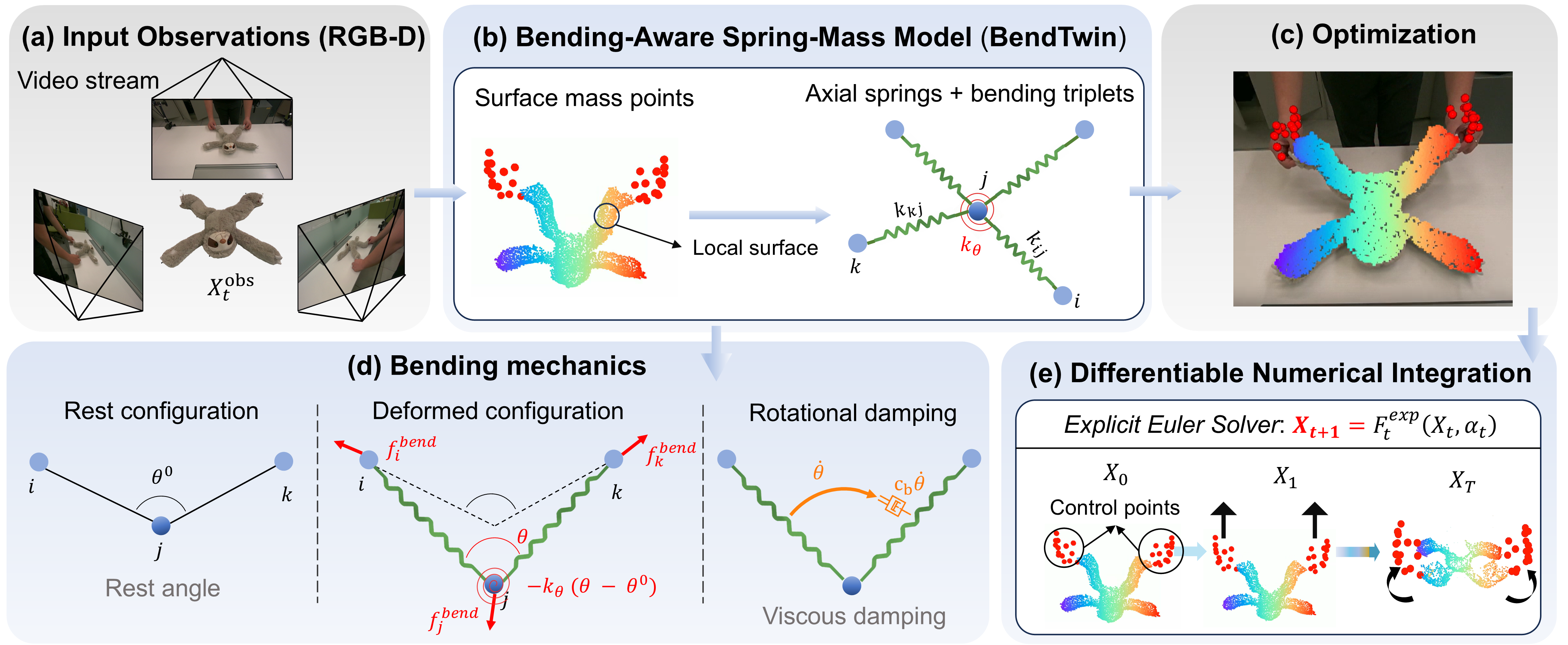}
  \caption{\model{} framework overview. (a) represents the 3D reconstruction from sparse-view RGB-D observations into a point-based surface discretization. (b) indicates that \model{} extends the spring--mass model with explicit bending triplets in addition to axial springs and (c) optimizes physical parameters using geometry and tracking losses. (d) visualizes the details of the proposed bending mechanism. (e) demonstrates differentiable explicit simulation of the bending-aware spring--mass system.
  }
  \label{fig:method_pipeline}
\end{figure}

We present \model{}, a bending-aware differentiable physics framework illustrated in Fig.~\ref{fig:method_pipeline}(a--e). 
We formulate the inverse problem of estimating mechanical parameters by matching simulated dynamics to partial 3D observations from videos. 
\model{} augments axial spring interactions with explicit bending stiffness and damping over surface triplets, adding bending constraints that improve the mechanical stability of sparse spring--mass systems. The resulting model preserves the simplicity of explicit differentiable simulation while reducing reliance on dense spring--mass systems.

\subsection{Problem Formulation}
\label{sec:formulation}

Given multi-view RGB-D videos of an object undergoing interaction, we aim to reconstruct a physically consistent digital twin by estimating its mechanical parameters using \model{}. As illustrated in Fig.~\ref{fig:method_pipeline}(a--e), the pipeline starts from sparse-view RGB-D observations, which are then discretized into a spring--mass system for inverse mechanical parameter estimation.

We discretize the object as a graph $\mathcal{G} = \{\mathcal{V}, \mathcal{E}, \mathcal{T}\}$, where $\mathcal{V}$ denotes a set of surface mass nodes connected by axial springs $\mathcal{E}$ and bending triplets $\mathcal{T}$. Each node $i\in\mathcal{V}$ has position $x_i^t\in\mathbb{R}^3$ and velocity $v_i^t\in\mathbb{R}^3$ at time $t$, and the system state at time $t$ is $X_t = \{x_i^t, v_i^t\}_{i\in\mathcal{V}}$.

Let $X_t^{\text{obs}}$ denote the partial 3D observation of the object at time $t$ as shown in Fig.~\ref{fig:method_pipeline}a (\eg a point cloud lifted from sparse-view RGB-D frames and completed using TRELLIS~\cite{11094803}, which is consistent with the pre-processing in PhysTwin~\cite{jiang2025phystwin}). Our objective is to estimate the physical parameters of the deformable object, which we collect into
\begin{equation}
\alpha = \{k_{ij}, \gamma_{ij}\}_{(i,j)\in \mathcal{E}} \cup \{k_{\theta,(i,j,k)}, c_{b,(i,j,k)} \}_{(i,j,k)\in\mathcal{T}} \cup \alpha_{\text{contact}},
\end{equation}
where $k_{ij}$ is the axial stiffness and $\gamma_{ij}$ is the dashpot damping coefficient. $k_{\theta,(i,j,k)}$ (see Fig.~\ref{fig:method_pipeline}b) penalizes deviations from the rest angle of triplet $(i,j,k)$, and $c_{b,(i,j,k)}$ specifies viscous bending damping proportional to the angular rate $\dot\theta_{(i, j, k)}$. Fig.~\ref{fig:method_pipeline}d illustrates the details of bending mechanics, \ie the rest angle of a triplet, the restoring forces induced by angular deviation, and the viscous bending damping. $\alpha_{\text{contact}}$ parameterize the contact model (\eg restitution coefficient and Coulomb friction coefficient for object--object and object--ground interactions). As shown in Fig.~\ref{fig:method_pipeline}c, our objective minimizes the discrepancy between simulated and observed motion:
\begin{equation}{
\resizebox{0.9\linewidth}{!}{$
\min_{\alpha,\,\mathcal{G}_t}\ \mathcal{L}(\hat X_t, X_t^{\text{obs}}) = \sum_{t=0}^{T}
\Big(
\mathcal{L}_{\text{geo}}(\hat X_t, X_t^{\text{obs}})
+
\mathcal{L}_{\text{trk}}(\hat X_t, X_t^{\text{obs}})
\Big)
\quad \text{s.t.}\quad
\hat X_{t+1} = F^{\text{exp}}_{\alpha,\,\mathcal{G}_t}(\hat X_t).$
    }}
\label{eq:inv_problem}
\end{equation}
where $\hat X_t$ denotes the simulated system state at time $t$. $\mathcal{L}_{\text{geo}}$ and $\mathcal{L}_{\text{trk}}$ measure the geometric and motion discrepancies between $\hat X_t$ and the partial observation $X_t^{\text{obs}}$ using Chamfer distance and point-wise tracking, respectively. 
To update the system dynamics, we employ a differentiable explicit Euler integrator $F^{\text{exp}}_{\alpha,\,\mathcal{G}_t}(\cdot)$ (see Fig.~\ref{fig:method_pipeline}e). This enables efficient forward simulation while maintaining numerical stability in contact-rich scenarios.

\subsection{Bending-Enhanced Spring--Mass Dynamics}
\label{sec:bending_model}

Naive spring--mass models (used in PhysTwin~\cite{jiang2025phystwin} and Spring-Gau~\cite{zhong2024reconstruction}) rely solely on axial springs to resist stretching and compression. While effective for modeling compliant deformation, axial-only interactions do not explicitly model bending mechanics. Consequently, bending-dominated surface deformations must be approximated indirectly through axial coupling, limiting the mechanical expressiveness of the system and reducing its ability to faithfully reproduce complex deformation behaviors.

To address these limitations, \model{} augments the spring--mass model with explicit bending stiffness defined directly over local surface configurations. Specifically, we introduce bending springs over node triplets $(i, j, k)$ to penalize angular deviations, allowing bending resistance to be modeled explicitly. These bending constraints complement axial springs and reduce under-constrained local deformation modes when the system contains fewer sampled nodes and connections. Meanwhile, \model{} preserves the simplicity of spring--mass systems while improving the mechanical stability of sparse discretizations.

We define the total internal potential energy $E_{\text{int}}(X)$ as
\begin{equation}
E_{\text{int}}(X) = \sum_{(i,j)\in \mathcal{E}} E_{\text{axial}}(i,j) \;+\; \sum_{(i,j,k)\in\mathcal{T}} E_{\text{bend}}(i,j,k).
\end{equation}
where $E_{\text{axial}}(i,j)$ denotes the standard axial spring energy, whose induced nodal forces are identical to those used in PhysTwin~\cite{jiang2025phystwin} and therefore omitted for brevity.
This paper only introduces the derivation of the additional forces introduced by the proposed bending energy $E_{\text{bend}}(i,j,k)$, which can be obtained by differentiating $E_{\text{bend}}$ with respect to node positions:
\begin{equation}
f^{\text{bend}} = -\nabla_x E_{\text{bend}}.
\end{equation}

\subsubsection{Bending Stiffness}
\label{sec:rot_stiff}

To model bending, we introduce bending springs over local triplets
$(i,j,k)\in\mathcal{T}$, where $j$ is the center node and $i,k$ are its neighbors.
The rest angle $\theta^0$ corresponds to the angle between edges $(i,j)$ and $(k,j)$
in $\mathcal{G}_0$. We define the bending energy as
\begin{equation}
E_{\text{bend}}(i,j,k)
=
\tfrac{1}{2}k_{\theta,(i,j,k)}(\theta-\theta^0)^2,
\label{eq:bend_energy}
\end{equation}
where $\theta$ denotes the current angle between the two edges.

Differentiating Eq.~\eqref{eq:bend_energy} yields restoring forces that act to reduce the angular deviation:
\begin{equation}
f^{\text{bend}}_{i} = -k_{\theta}(\theta-\theta^0)\,\mathbf{d}_{i}^{(i,j,k)},
\qquad
f^{\text{bend}}_{k} = -k_{\theta}(\theta-\theta^0)\,\mathbf{d}_{k}^{(i,j,k)}.
\end{equation}
where $\mathbf{d}_{i}^{(i,j,k)}$ and $\mathbf{d}_{k}^{(i,j,k)}$ are unit-scaled
directional vectors determined by the local geometry of the $(i, j, k)$. These directions ensure torque-consistent application of force and are derived in the Appendix. By Newton's third law (force balance within the $(i, j, k)$), the force on $j$ is
\begin{equation}
f^{\text{bend}}_{j} = -\left(f^{\text{bend}}_{i} + f^{\text{bend}}_{k}\right)
= k_{\theta}(\theta-\theta^0)\left(\mathbf{d}_{i}^{(i,j,k)}+\mathbf{d}_{k}^{(i,j,k)}\right).
\end{equation}

\subsubsection{Bending Damping}
\label{sec:rot_damp}

In addition to elastic bending stiffness, we introduce bending damping to dissipate energy
associated with angular motion to avoid the oscillations of the system caused by the extra bending stiffness. For each bending triplet $(i,j,k)$, we define an angular rate
$\dot{\theta}_{(i,j,k)}$ that measures the relative angular motion of edges $(i,j)$ and $(k,j)$. We model bending damping as a viscous torque proportional to this angular rate, with damping
coefficient $c_{b,(i,j,k)}$. The resulting nodal damping forces are
\begin{equation}
f^{\text{rot}}_{i} = -c_{b}\,\dot{\theta}_{(i,j,k)}\,\mathbf{d}_{i}^{(i,j,k)},
\qquad
f^{\text{rot}}_{k} = -c_{b}\,\dot{\theta}_{(i,j,k)}\,\mathbf{d}_{k}^{(i,j,k)}.
\end{equation}
The equivalent damping force introduced on the center node $j$ is
\begin{equation}
f^{\text{rot}}_{j} = -\left(f^{\text{rot}}_{i} + f^{\text{rot}}_{k}\right)
= c_{b}\,\dot{\theta}_{(i,j,k)}\left(\mathbf{d}_{i}^{(i,j,k)}+\mathbf{d}_{k}^{(i,j,k)}\right).
\end{equation}
The force directions $\mathbf{d}_{i}^{(i,j,k)}$ and $\mathbf{d}_{k}^{(i,j,k)}$ are identical to those used for elastic bending forces. The explicit computation of $\dot{\theta}_{(i,j,k)}$ from nodal velocities is provided in the Appendix.

\subsubsection{Efficient Triplet Construction for High-Valence Nodes}
\label{sec:triplets}
If node $j$ has $N$ neighbors, enumerating all neighbor pairs yields $\mathcal{O}(N^2)$ triplets, which is expensive and often introduces redundant stiffness in the model. Instead, we construct $\mathcal{O}(N)$ triplets by selecting a reference neighbor $r(j)\in\mathcal{N}(j)$ and forming triplets $(r(j),j,k)$ for all $k\in\mathcal{N}(j)\setminus\{r(j)\}$ (or a random subset). This preserves bending expressivity while keeping the model scalable.

Given the axial springs, bending triplets, and external interactions, the total force acting on node $j$ is computed as
\begin{equation}
f_j^{\text{tot}}
=
\underbrace{
\sum_{i\in\mathcal{N}_{\mathcal{E}}(j)}
\big(
f^{\text{axial}}_{j,i} + f^{\text{dash}}_{j,i}
\big)
}_{\text{axial spring and damping}}
\;+\;
\underbrace{
\sum_{(i,j,k)\in\mathcal{T}_j}
\big(
f^{\text{bend}}_{j,(i,j,k)} + f^{\text{rot}}_{j,(i,j,k)}
\big)
}_{\text{bending stiffness and bending damping}}
\;+\;
\underbrace{
f^{\text{ext}}_j
}_{\text{external forces}}.
\label{eq:total_force_node_j}
\end{equation}

\section{Experiments}
\label{sec:experiments}

In this section, we evaluate \model{} to assess the impact of bending stiffness on mechanical fidelity and simulation stability. Specifically, we examine (1) whether bending-aware constraints improve object reconstruction \& re-simulation and future prediction over the axial-only PhysTwin baseline; and (2) whether bending constraints preserve accuracy and stability as the physical system is progressively downsampled.

\subsection{Experimental Settings}
\label{sec:exp_settings}

\textbf{Dataset.} We use the non-cloth deformable object sequences from the PhysTwin benchmark~\cite{jiang2025phystwin}. The revised evaluation excludes fabric-like sheet sequences because cloth is highly compliant and exhibits little intrinsic resistance to bending or shearing, whereas our bending-aware model assumes objects with sufficient structural support for stable bending constraints. Following the original protocol, each sequence is split into training and testing frames with a 7:3 ratio. The training split is used for the reconstruction \& re-simulation, while the test split evaluates the future prediction.

\noindent \textbf{Baselines.} The primary comparison is against PhysTwin~\cite{jiang2025phystwin}, which adopts an axial spring--mass formulation. This comparison directly evaluates whether incorporating bending constraints improves accuracy and enables the assessment of mechanical stability in sparse physical systems.

\noindent \textbf{Evaluation metrics.}  To evaluate reconstruction and prediction accuracy, we report 3D metrics including Chamfer Distance (CD) and tracking error, and 2D rendering metrics including PSNR, SSIM~\cite{wang2004image}, LPIPS~\cite{zhang2018unreasonable}, and silhouette IoU computed from 3D Gaussian Splatting renderings.

\noindent \textbf{Implementation details.} Our implementation is based on PyTorch and Warp for GPU-accelerated spring--mass simulation. All experiments are performed on one NVIDIA RTX 5090 GPU. We follow PhysTwin~\cite{jiang2025phystwin} for 3D reconstruction. Parameters $\alpha$ are optimized via Adam with lr $10^{-3}$ to minimize $\mathcal{L}_{\text{geo}} + \mathcal{L}_{\text{trk}}$ with equal weights. For the active non-cloth evaluation, all simulations use the differentiable explicit Euler rollout $F^{\text{exp}}$ with sub-step $\Delta t = 5\times10^{-5}$ s, $M_t \approx 667$ sub-steps per frame, and maximum axial stiffness $k_{ij}=10^5$. Collision parameters $\alpha_{\text{contact}}$ use restitution 0.5/0.7 and friction 0.3.

\subsection{Results}

\noindent \textbf{Reconstruction \& re-simulation.} We first evaluate reconstruction \& re-simulation on the non-cloth object set using the explicit Euler simulator for all methods. The quantitative results are reported in Table~\ref{tab:main_resim_future}. Compared with the axial-only PhysTwin baseline, \model{} reduces CD from 0.0058 to 0.0047 (\eg \textbf{19.0\%} improvement), and reduces tracking error from 0.0091 to 0.0078 (\eg \textbf{14.3\%} improvement). These gains indicate that the added bending stiffness and damping provide useful local shape constraints beyond pairwise axial springs, allowing the physical graph to better preserve object structure during re-simulation.

The improvement on geometric accuracy is reflected in the rendered observations. \model{} improves PSNR from 28.429 to 28.605, a gain of \textbf{0.18 dB}, raises IoU from 79.5 to 80.6, and reduces LPIPS from 0.022 to 0.021. Overall, the results show that explicit bending constraints make the reconstructed dynamics more geometrically accurate while maintaining comparable visual quality for deformable objects.

\noindent \textbf{Future prediction.} We further evaluate future prediction by optimizing physical parameters on the training portion of each sequence and then simulating forward to unseen frames. As shown in Table~\ref{tab:main_resim_future}, \model{} also improves long-horizon prediction over PhysTwin. CD decreases from 0.0095 to 0.0079 (\eg \textbf{16.8\%} reduction), and tracking error decreases from 0.0166 to 0.0153 (\eg \textbf{7.8\%} reduction). These results suggest that bending-aware constraints reduce the accumulation of geometric drift when the optimized model is rolled out beyond the observed training frames. The rendering metrics follow the same trend, with \model{} improving IoU from 68.7 to 71.5 and PSNR from 26.412 to 26.755, corresponding to gains of \textbf{2.8} percentage points and \textbf{0.34 dB}, respectively.

\begin{table*}[t]
\centering
\caption{Quantitative results on Reconstruction \& Re-simulation and Future Prediction.}
\label{tab:main_resim_future}
\resizebox{\textwidth}{!}{
\begin{tabular}{l|cccccc|cccccc}
\toprule
& \multicolumn{6}{c|}{Reconstruction \& Re-simulation} & \multicolumn{6}{c}{Future Prediction} \\
\cline{2-13}
Model
& \rule[-0.8ex]{0pt}{2.8ex}CD $\downarrow$ & Track $\downarrow$ & IoU \% $\uparrow$ & PSNR $\uparrow$ & SSIM $\uparrow$ & LPIPS $\downarrow$
& CD $\downarrow$ & Track $\downarrow$ & IoU \% $\uparrow$ & PSNR $\uparrow$ & SSIM $\uparrow$ & LPIPS $\downarrow$ \\
\midrule
PhysTwin
& 0.0058 & 0.0091 & 79.5 & 28.429 & 0.963 & 0.022
& 0.0095 & 0.0166 & 68.7 & 26.412 & 0.957 & 0.036 \\
\model{} (Ours)
& \textbf{0.0047} & \textbf{0.0078} & \textbf{80.6} & \textbf{28.605} & \textbf{0.963} & \textbf{0.021}
& \textbf{0.0079} & \textbf{0.0153} & \textbf{71.5} & \textbf{26.755} & \textbf{0.957} & \textbf{0.035} \\
\bottomrule
\end{tabular}}
\end{table*}

\begin{figure}[tb]
  \centering
  \includegraphics[width=1.0\linewidth]{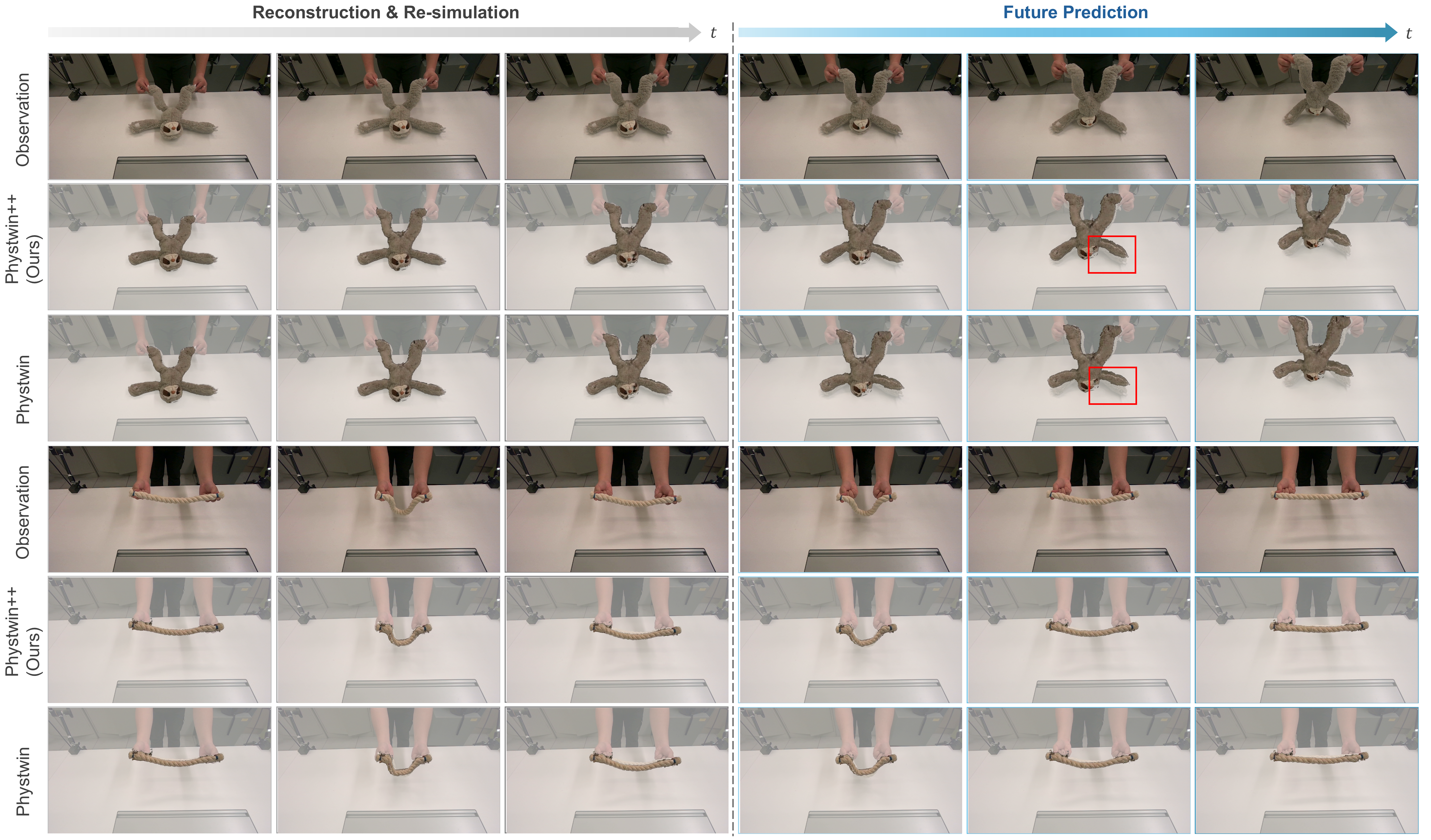}
  \caption{
  The qualitative results of \model{} and the comparison method on the reconstruction\&re-simulation and future predictions of the object deformation. The qualitative results show that our method is closer to the true values than other methods, with red boxes highlighting regions where \model{} better captures object deformation than PhysTwin.
  }
  \label{fig:results}
\end{figure}

These results indicate that explicitly modeling bending stiffness leads to more physically consistent motion propagation over time. By capturing higher-order mechanical behavior beyond axial deformation, the proposed model produces more accurate geometric deformation and dynamic trajectories. Qualitative results in Fig.~\ref{fig:results} further demonstrate that \model{} better matches the ground-truth deformation patterns during long-horizon prediction.

\subsection{Ablation Studies}

\paragraph{Downsampled physical system.}
We evaluate whether providing extra bending constraints maintains stable and accurate dynamics as the physical graph is downsampled. For each sampling ratio $r$, both PhysTwin and \model{} downsample the full physical graph at the same ratio $r$, ensuring a matched graph. The only difference between the two systems is the presence of the proposed bending springs. A smaller $r$ retains fewer mass points (\eg progressively compressing the physical system) for testing whether bending constraints preserve stability where axial springs alone become under-constrained.

\begin{table*}[t]
\centering
\caption{Robustness to physical-graph downsampling. Both PhysTwin and \model{} retain interior and exterior nodes and are downsampled at the same ratio $r$; the only difference is the proposed bending springs.}
\label{tab:ablation_sampling}
\resizebox{\textwidth}{!}{
\begin{tabular}{c l|cccccc|cccccc}
\toprule
\multirow{2}{*}{Sampling ratio $r$}
& \multirow{2}{*}{Model}
& \multicolumn{6}{c|}{Reconstruction \& Re-simulation}
& \multicolumn{6}{c}{Future Prediction} \\
\cline{3-14}
& & \rule[-0.8ex]{0pt}{2.8ex}CD $\downarrow$ & Track $\downarrow$ & IoU \% $\uparrow$ & PSNR $\uparrow$ & SSIM $\uparrow$ & LPIPS $\downarrow$
& CD $\downarrow$ & Track $\downarrow$ & IoU \% $\uparrow$ & PSNR $\uparrow$ & SSIM $\uparrow$ & LPIPS $\downarrow$ \\
\midrule
\multirow{2}{*}{0.05}
& PhysTwin
& 0.0139 & 0.0175 & 72.2 & 26.926 & 0.958 & 0.0309
& 0.0216 & 0.0241 & 60.8 & 25.276 & 0.953 & 0.0476 \\
& \model{} (Ours)
& \textbf{0.0102} & \textbf{0.0130} & \textbf{77.2} & \textbf{27.996} & \textbf{0.961} & \textbf{0.0262}
& \textbf{0.0176} & \textbf{0.0202} & \textbf{67.8} & \textbf{26.417} & \textbf{0.955} & \textbf{0.0411} \\
\midrule
\multirow{2}{*}{0.1}
& PhysTwin
& 0.0088 & 0.0120 & 78.2 & 28.105 & 0.962 & 0.0245
& 0.0150 & 0.0198 & 66.6 & 26.116 & 0.955 & 0.0409 \\
& \model{} (Ours)
& \textbf{0.0079} & \textbf{0.0107} & \textbf{79.7} & \textbf{28.552} & \textbf{0.963} & \textbf{0.0222}
& \textbf{0.0139} & \textbf{0.0181} & \textbf{69.1} & \textbf{26.620} & \textbf{0.957} & \textbf{0.0362} \\
\bottomrule
\end{tabular}}
\end{table*}

Table~\ref{tab:ablation_sampling} shows that \model{} consistently outperforms the axial-only PhysTwin across all metrics at both downsampling ratios, and that its advantage grows substantially as the physical graph becomes sparser. At $r=0.1$, \model{} reduces reconstruction CD by \textbf{9.6\%} and tracking error by \textbf{10.8\%}, while improving PSNR by \textbf{0.45~dB} and IoU by \textbf{1.6} percentage points. When the graph is halved again to $r=0.05$, where the axial-only formulation becomes severely under-constrained, the improvement roughly \emph{doubles}: CD is reduced by \textbf{26.5\%} and tracking error by \textbf{25.6\%}, with PSNR improving by \textbf{1.07~dB} and IoU by \textbf{4.9} percentage points.

The same trend holds for future prediction, where the gains grow from \textbf{7.2\%} CD and \textbf{8.5\%} tracking-error reduction at $r=0.1$ to \textbf{18.3\%} and \textbf{16.3\%} at $r=0.05$, with PSNR gains increasing from \textbf{0.50~dB} to \textbf{1.14~dB} and IoU gains from \textbf{2.5} to \textbf{7.0} percentage points. This widening of the margin demonstrates that as connectivity is progressively removed, the axial-only system loses the constraints needed to preserve local shape and degrades sharply. However, the bending stiffness in \model{} continues to penalize non-physical angular deviations and keeps the sparse spring--mass system stable.

\paragraph{Removal of interior nodes.}
PhysTwin discretizes each volumetric object into exterior surface nodes and interior nodes. Although the interior nodes are unobserved and excluded from evaluation, the axial springs incident to them provide internal mechanical support and help prevent the surface from collapsing inward during interaction. We argue that \model{} no longer requires this support because its bending term directly penalizes changes in local surface angles, thereby resisting the non-physical surface distortions that the interior axial springs are intended to suppress. A volumetric object may therefore be represented as a surface-only physical system. We demonstrate it with an ablation at full sampling density ($r=1.0$), removing all interior nodes from \model{} and simulating only the surface nodes. We compare this surface-only configuration with the axial-only PhysTwin baseline, which retains its complete interior discretization. All metrics are evaluated on surface points for both methods, ensuring that the evaluation support is identical.

\begin{table*}[t]
\centering
\caption{Ablation of interior nodes at full sampling density ($r=1.0$). \model{} is simulated using surface nodes only, whereas PhysTwin retains its complete interior discretization. All metrics are evaluated on surface points.}
\label{tab:ablation_interior_dense}
\resizebox{\textwidth}{!}{
\begin{tabular}{l|c|cccccc|cccccc}
\toprule
\multirow{2}{*}{Configuration}
& \multirow{2}{*}{Interior nodes}
& \multicolumn{6}{c|}{Reconstruction \& Re-simulation}
& \multicolumn{6}{c}{Future Prediction} \\
\cline{3-14}
& & \rule[-0.8ex]{0pt}{2.8ex}CD $\downarrow$ & Track $\downarrow$ & IoU \% $\uparrow$ & PSNR $\uparrow$ & SSIM $\uparrow$ & LPIPS $\downarrow$
& CD $\downarrow$ & Track $\downarrow$ & IoU \% $\uparrow$ & PSNR $\uparrow$ & SSIM $\uparrow$ & LPIPS $\downarrow$ \\
\midrule
PhysTwin
& w/
& 0.0058 & 0.0091 & 79.5 & 28.429 & 0.963 & 0.0225
& 0.0095 & \textbf{0.0166} & 68.7 & 26.412 & 0.957 & 0.0357 \\
\model{} (Ours)
& w/o
& \textbf{0.0054} & \textbf{0.0088} & \textbf{80.6} & \textbf{28.487} & 0.963 & \textbf{0.0220}
& \textbf{0.0090} & 0.0167 & \textbf{72.0} & \textbf{26.982} & \textbf{0.959} & \textbf{0.0333} \\
\bottomrule
\end{tabular}}
\end{table*}

As reported in Table~\ref{tab:ablation_interior_dense}, removing the entire interior discretization does not degrade the surface-level accuracy of \model{}. For reconstruction and re-simulation, the surface-only \model{} reduces CD by approximately \textbf{6.9\%} and tracking error by \textbf{3.3\%}, while increasing IoU by \textbf{1.1} percentage points. The gains are more pronounced for future prediction, \ie CD and LPIPS decrease by approximately \textbf{5.3\%} and \textbf{6.7\%}, respectively, while PSNR increases by \textbf{0.570~dB} and IoU by \textbf{3.3} percentage points. Bending constraints are therefore sufficient to replace the mechanical role of interior nodes. A pure surface system reproduces volumetric deformation as faithfully as an axial-only system that supports its interior explicitly, while simulating fewer mass points.

\section{Related Works}
\label{sec:related_work}

\subsection{3D Reconstruction with Dynamics.}

Dynamic reconstruction goes beyond multi-view geometric consistency and additionally requires temporal coherence modeling to capture long-term changes over time. Classical approaches typically rely on inter-frame correspondence estimation, such as feature-based matching ~\cite{lowe2004sift,mur2017orbslam2} to perform tracking and reconstruction across frames. However, these pipelines often struggle to produce dense reconstructions due to sparse keypoint reliance and tracking drift over long sequences. 

Recent advances in neural scene representations, particularly Neural Radiance Fields (NeRF)~\cite{mildenhall2021nerf}, have introduced implicit volumetric representations capable of synthesizing high-fidelity novel views. Extensions to dynamic settings ~\cite{pumarola2021dnerf, park2021nerfies}, incorporate temporal deformation fields to model non-rigid scene evolution. More recently, 3D Gaussian Splatting~\cite{kerbl20233d} has emerged as an explicit and efficient 3D representation enabling real-time, high-quality rendering. Dynamic extensions such as 4D Gaussian representations~\cite{wu20244dgs, yang2024gaussianflow} tightly integrate visual tracking, spatio-temporal representation learning, and differentiable rendering, achieving temporally consistent dynamic scene reconstruction with improved efficiency and fidelity.

Despite these advances, current reconstruction methods remain largely geometry- and appearance-driven. The reconstructed objects typically lack integrated physical priors, which limits their ability to predict future states or support physically grounded interaction. Without embedding physics-based constraints or dynamics modeling, reconstructed 4D scenes function primarily as descriptive representations rather than interactive prediction or action simulation.

\subsection{Physical Simulation and Differentiable Physics.}

Classical physical simulation in computer graphics relies on established numerical formulations such as Lagrangian mechanics~\cite{zefran2005lagrangian}, finite element methods~\cite{sifakis2012fem, 5980327}, material point methods (MPM)~\cite{stomakhin2013material, jiang2016material, hu2018moving}, and position-based dynamics~\cite{muller2007position, macklin2016xpbd}. These approaches have enabled realistic simulation of rigid and deformable objects across a wide range of applications. However, they typically assume that material parameters are manually specified or calibrated offline, limiting their applicability when physical properties are unknown.

Differentiable physics aims to bridge this gap by making complex dynamic systems differentiable, allowing unknown physical parameters to be optimized from observations using gradient-based methods. A number of differentiable simulators have been developed for rigid-body dynamics~\cite{peres_2018, degrave2019differentiable, macklin2020primal, ijcai2019p869} and soft-body dynamics~\cite{geilinger2020add, hahn2019real2sim, hu2019chainqueen, hu2019taichi, qian2020diffphy, liang2019differentiable, du2021diffpd, 2026arXiv260615015Y}. While these methods enable parameter optimization by fitting simulated dynamics to observations, they do not reconstruct a complete digital twin that simultaneously recovers geometry, material behavior, and appearance from sparse visual inputs.

\subsection{Physics-Driven 3D Reconstruction.}

Recent works incorporate continuum mechanics into neural scene representations such as NeRF and 3D Gaussian Splatting to enable dynamic deformation and interaction~\cite{xie2024physgaussian, zhang2024physdreamer, feng2024pie, xie2025physanimator}. However, these approaches typically assume predefined material properties and focus on visually plausible motion rather than recovering true object dynamics. Differentiable physics  addresses this limitation by enabling optimization of physical parameters directly from RGB observations~\cite{li2023pacnerf, gao2025seeing, 2026arXiv260507687J}, but such methods often operate on implicit geometry and do not explicitly reconstruct physically consistent 3D structure.

A more closely related line of work constructs physical digital twins using spring--mass models grounded in reconstructed geometry. Spring-Gaussian~\cite{zhong2024reconstruction} couples 3D Gaussian splatting with elastic simulation, and PhysTwin~\cite{jiang2025phystwin} reconstructs deformable objects from sparse-view RGB-D videos and differentiably optimizes a spring--mass system to reproduce observed interactions. While flexible and scalable, these axial-spring formulations lack explicit modeling of more complex mechanical behaviors, such as bending-dominated deformation, which motivates our proposed method.
\section{Conclusion}

We presented \model{}, a bending-aware differentiable spring--mass framework for physics-informed reconstruction and future prediction of deformable-object dynamics from sparse-view RGB-D videos. By augmenting axial spring interactions with explicit bending stiffness and damping over local surface triplets, \model{} penalizes deviations from rest angles and improves mechanical expressiveness while preserving the simplicity and differentiability of spring--mass models. 
\model{} demonstrates competitive performance across both dense and sparse spring–mass systems.
Overall, \model{} provides an efficient and mechanically expressive foundation for deformable-object digital twins in robotic manipulation, planning, and physical interaction.

%
%


\newpage

\appendix

\section{Derivation of Bending Forces and Bending Damping}
\label{sec:appendix_bend_rot}

In this section, we provide the derivations for the bending restoring forces and the bending
damping forces induced by a bending triplet $(i,j,k)$.
Following the main paper, we adopt the compact notation
$f_i^{\text{bend}} = -k_{\theta}(\theta-\theta^0)\mathbf{d}_i^{(i,j,k)}$ and
$f_i^{\text{rot}} = -c_b \dot{\theta}_{(i,j,k)}\mathbf{d}_i^{(i,j,k)}$, where the explicit
expressions of $\mathbf{d}_i^{(i,j,k)}$ and $\mathbf{d}_k^{(i,j,k)}$ are derived below.

\subsection{Notation}
For a bending triplet $(i,j,k)$ with center node $j$, define
\begin{equation}
\resizebox{0.9\linewidth}{!}{$
u = x_i - x_j,\qquad
v = x_k - x_j,\qquad
\ell_u=\|u\|,\qquad
\ell_v=\|v\|,\qquad
\tilde u = \frac{u}{\ell_u},\qquad
\tilde v = \frac{v}{\ell_v}
$}
\end{equation}
The included angle is computed from
\begin{equation}
c = \tilde u^\top \tilde v = \cos\theta,\qquad
s = \sqrt{1-c^2} = \sin\theta,\qquad
\theta = \arccos(c),
\end{equation}
and $\theta^0$ denotes the rest angle in the canonical configuration.
The bending energy is
\begin{equation}
E_{\text{bend}}(i,j,k) = \tfrac{1}{2}k_\theta(\theta-\theta^0)^2,
\label{eq:appendix_bend_energy}
\end{equation}
with force convention $f=-\nabla_x E_{\text{bend}}$.

\subsection{Derivation of Bending Forces}
Differentiating \eqref{eq:appendix_bend_energy} gives
\begin{equation}
\nabla_x E_{\text{bend}}
=
k_\theta(\theta-\theta^0)\nabla_x \theta
=
-\frac{k_\theta(\theta-\theta^0)}{s}\nabla_x c,
\end{equation}
where we used $\nabla_x\theta = -\frac{1}{s}\nabla_x c$.
A standard calculation yields
\begin{equation}\label{eq:derivation_dir}
\frac{\partial c}{\partial u}
=
\frac{1}{\ell_u}\left(\tilde v - c\tilde u\right),
\qquad
\frac{\partial c}{\partial v}
=
\frac{1}{\ell_v}\left(\tilde u - c\tilde v\right),
\end{equation}
and since $u=x_i-x_j$ and $v=x_k-x_j$,
\begin{equation}
\frac{\partial c}{\partial x_i} = \frac{\partial c}{\partial u},
\qquad
\frac{\partial c}{\partial x_k} = \frac{\partial c}{\partial v},
\qquad
\frac{\partial c}{\partial x_j} = -\frac{\partial c}{\partial u}-\frac{\partial c}{\partial v}.
\end{equation}
Combining the above with $f=-\nabla_x E_{\text{bend}}$ yields the nodal forces
\begin{align}
f_i^{\text{bend}}
&=
k_\theta(\theta-\theta^0)\frac{1}{\ell_u s}\left(\tilde v - c\tilde u\right),
\label{eq:appendix_fi_bend}
\\
f_k^{\text{bend}}
&=
k_\theta(\theta-\theta^0)\frac{1}{\ell_v s}\left(\tilde u - c\tilde v\right),
\label{eq:appendix_fk_bend}
\\
f_j^{\text{bend}}
&=
-\left(f_i^{\text{bend}}+f_k^{\text{bend}}\right).
\label{eq:appendix_fj_bend}
\end{align}
Accordingly, the directional vectors used in the main paper can be identified as
\begin{equation}
\mathbf{d}_i^{(i,j,k)} = \frac{1}{\ell_u s}\left(\tilde v - c\tilde u\right),
\qquad
\mathbf{d}_k^{(i,j,k)} = \frac{1}{\ell_v s}\left(\tilde u - c\tilde v\right).
\label{eq:appendix_dirvecs}
\end{equation}

\subsection{Bending Damping}
We model bending damping as a viscous torque proportional to the angular rate:
\begin{equation}
\tau = -c_b \dot\theta_{(i,j,k)}.
\end{equation}
To compute $\dot\theta_{(i,j,k)}$, define the unit normal and tangential directions
\begin{equation}
n = \frac{\tilde u \times \tilde v}{\|\tilde u \times \tilde v\|},\qquad
t_u = n\times \tilde u,\qquad
t_v = n\times \tilde v.
\end{equation}
Let node velocities be $v_i, v_j, v_k$. A consistent first-order approximation is
\begin{equation}
\dot\theta_{(i,j,k)}
=
\frac{t_u^\top (v_i - v_j)}{\ell_u}
-
\frac{t_v^\top (v_k - v_j)}{\ell_v}.
\label{eq:appendix_theta_dot}
\end{equation}
The derivation of the equivalent nodal damping forces follows the same step as Equation~\ref{eq:derivation_dir}, therefore taking the same geometric directions as the elastic bending forces:
\begin{align}
f_i^{\text{rot}}
&=
c_b\dot\theta_{(i,j,k)}\,\mathbf{d}_i^{(i,j,k)},
\\
f_k^{\text{rot}}
&=
c_b\dot\theta_{(i,j,k)}\,\mathbf{d}_k^{(i,j,k)},
\\
f_j^{\text{rot}}
&=
-\left(f_i^{\text{rot}}+f_k^{\text{rot}}\right),
\end{align}
where $\mathbf{d}_i^{(i,j,k)}$ and $\mathbf{d}_k^{(i,j,k)}$ are given in
\eqref{eq:appendix_dirvecs}.

\paragraph{Numerical considerations.}
When $\theta\approx 0$ or $\theta\approx \pi$, $s=\sin\theta$ becomes small.
We clip $c$ to $[-1+\varepsilon,\,1-\varepsilon]$ and use $s=\sqrt{\max(1-c^2,\varepsilon_s)}$.
If $\|\tilde u\times \tilde v\|$ is below a threshold, we skip the damping term for that triplet.

\end{document}